\documentclass[11pt]{article}

\usepackage[T1]{fontenc}
\usepackage[utf8]{inputenc}
\DeclareUnicodeCharacter{202F}{ }   
\DeclareUnicodeCharacter{2248}{\approx}
\DeclareUnicodeCharacter{2013}{--}
\DeclareUnicodeCharacter{2014}{---}

\usepackage[a4paper,margin=1in]{geometry}
\usepackage{microtype}

\usepackage{amsmath,amssymb,amsthm}

\usepackage{graphicx}
\usepackage{booktabs}
\usepackage{tabularx}
\usepackage{multirow}
\usepackage{adjustbox}
\usepackage{subcaption}

\usepackage{enumitem}
\usepackage{xspace}
\usepackage{comment}
\usepackage{xcolor}

\usepackage{algorithm}
\usepackage{algorithmicx}
\usepackage{algpseudocode}

\usepackage{tikz}
\usetikzlibrary{positioning,arrows.meta,shapes,calc}
\usepackage{pgfplots}
\pgfplotsset{compat=1.17}
\usepackage[numbers,sort&compress]{natbib}

\usepackage{placeins}
\usepackage{float}

\usepackage[colorlinks=true,allcolors=blue]{hyperref}

\usepackage{authblk}

\newcommand{\keywords}[1]{\par\noindent\textbf{Keywords:} #1\par}

\title{Silent Collapse:Trajectory-Level Precursors in Recursive Learning}

\author[1,2]{Zhipeng Zhang}
\affil[1]{China Mobile Research Institute, Beijing 100053, China}
\affil[2]{China Mobile GBA (Greater Bay Area) Innovation Institute, Guangzhou 510656, China}

\date{} 

\begin{document}
\maketitle

\begin{center}
\textbf{Corresponding author:} Zhipeng Zhang (\texttt{zhangzhipeng@chinamobile.com})
\end{center}

\begin{abstract}

Recursive learning — where models iteratively learn from feedback whose reliability cannot be fully verified locally — is increasingly common in modern AI systems. However, standard endpoint validation metrics such as loss, perplexity, and accuracy may fail to reflect the true internal dynamical state of the learning system. Here we identify a phenomenon we call \textit{silent collapse}: a regime in recursive learning where latent geometric contraction and progressive loss of recoverability emerge before visible endpoint degradation.

We show that silent collapse follows an ordered dynamical sequence rather than an abrupt failure event. Across independent recursive trajectories, hidden contraction — measured through anchor entropy and representation dynamics — consistently precedes visible collapse. This hidden regime is accompanied by progressively weakened recoverability: later-stage checkpoints require substantially stronger intervention to restore stable learning dynamics under identical weak recovery conditions.

More importantly, we find that endpoint observables may become non-identifiable with respect to latent internal state. Checkpoints with nearly identical validation perplexity can occupy substantially separated latent geometry states and exhibit strongly different recovery dynamics under the same intervention protocol. Trajectory-level geometry analysis further reveals that recursive open-loop learning follows a continuous latent-state evolution while observable validation metrics become non-monotonic, producing self-intersections in observable space without corresponding intersections in latent geometry space.

Based on these trajectory-level precursors, we further explore an adaptive regulation strategy (Monitor--Trust--Regulator, MTR) that modulates recursive pressure using internal trajectory signals. Entropy-regulated trajectories remain substantially closer to an approximately injective mapping between latent geometry and observable behavior. Together, our results show that observable equivalence does not imply latent-state equivalence in recursive learning systems and that trajectory-level observables provide critical information unavailable from standard endpoint evaluation alone.

\end{abstract}


\section{Introduction}

Modern artificial intelligence systems increasingly learn iteratively
from feedback whose reliability cannot be fully verified from individual
observations alone. A language model fine-tunes on synthetic text
generated by an earlier version of itself
\citep{radford2019language,brown2020language}. A reinforcement learning
agent improves using rewards that may be systematically biased
\citep{christiano2017preferences,amodei2016concrete}. A self-supervised
classifier refines its predictions using pseudo-labels that become
increasingly confident — and increasingly wrong. More broadly, similar
feedback structures arise in AI-mediated scientific and decision-making
pipelines, including astronomy, medical prediction systems, climate
inference, and autonomous discovery workflows, where inferred quantities
may recursively influence future interpretation, selection, or optimization.
In all these cases, each individual training or inference signal may
appear locally plausible, yet the overall learning trajectory may drift
toward a degenerate state.

We define \textit{silent collapse} as a dynamical regime in which
iterative learning systems undergo progressive internal degeneration
— contraction of predictive entropy and representational dynamics,
and, in some settings, erosion of support statistics — while
endpoint validation metrics remain stable or improve.  Crucially,
this hidden degeneration is accompanied by a progressive loss of
recoverability: the deeper the system enters silent collapse, the
harder it becomes to restore healthy dynamics with a fixed budget
of real data.

We identify trajectory-level precursors that reliably emerge before
visible degradation, including robust contraction of anchor entropy
and representation drift, and (in some settings) erosion of support
statistics. These precursors manifest before any detectable change
in validation metrics, providing a practical early warning window.

A related phenomenon, \textit{model collapse}, was recently
demonstrated by \citet{shumailov2024ai} in recursively trained
generative models \citep{gerstgrasser2024model}. Their work showed
that models inevitably lose tail information and converge to delta
functions when trained on their own outputs. However, their analysis
left two critically important questions unanswered: is collapse
predictable before standard metrics degrade, and can it be prevented
without access to pristine real data? Here we show that recursive
degradation is both predictable before visible failure and partially
preventable without continued access to pristine real data. Beyond
providing a method for early warning and intervention, our results
reveal a more fundamental limitation: standard validation metrics alone
are insufficient to determine whether iterative learning dynamics
remain stable.

\begin{figure*}[t]
\centering
\includegraphics[width=\textwidth]{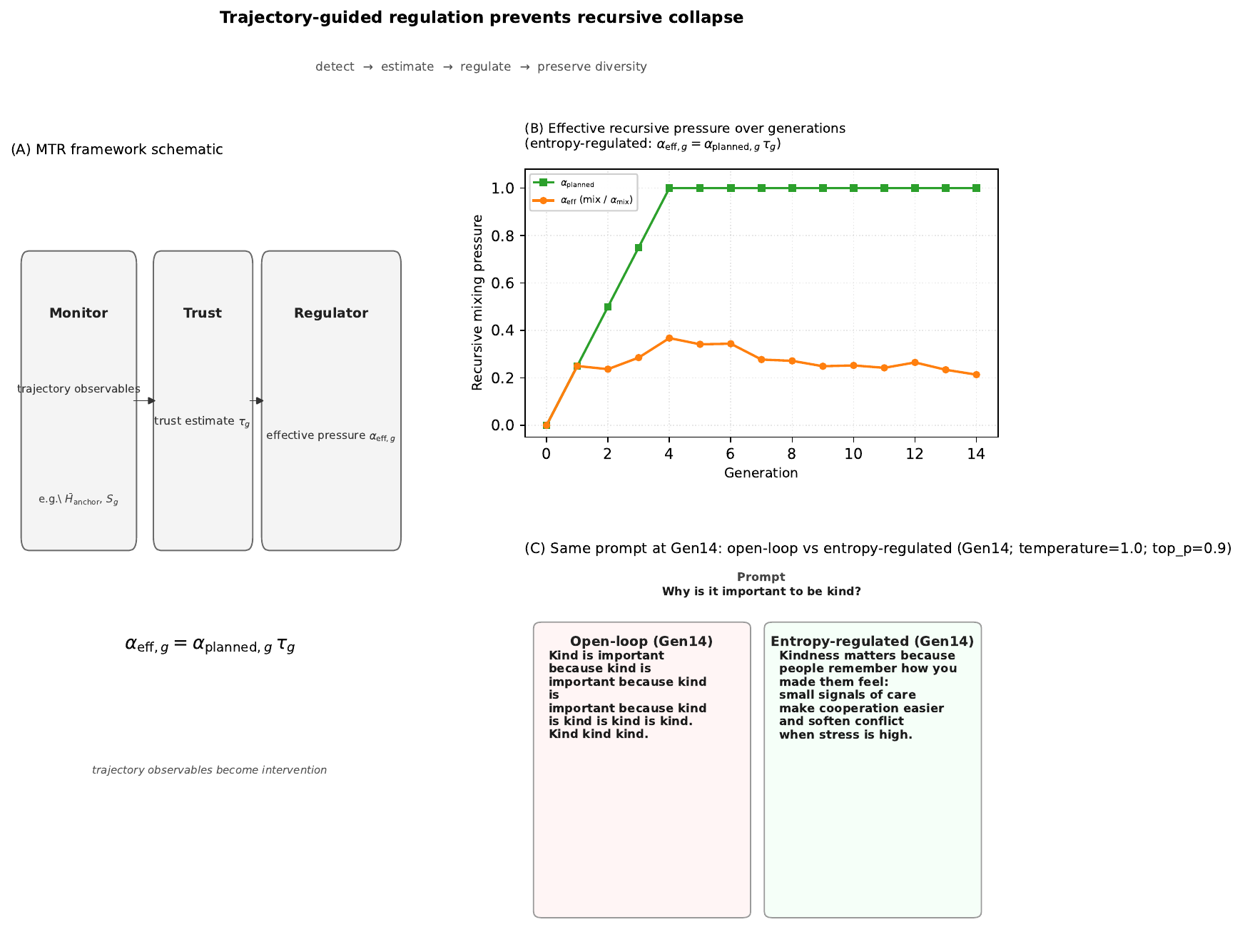}
\caption{\textbf{Trajectory-guided regulation mitigates recursive collapse under persistent recursive pressure.}
(A) MTR framework schematic. (B) Trajectory-guided regulation dynamically suppresses recursive synthetic exposure as anchor entropy contracts.
(C) Qualitative comparison between open-loop and MTR.}
\label{fig:mtr_framework}
\end{figure*}

Thus, silent collapse reframes recursive degradation as an
observability problem: the health of an iterative learning system
cannot be inferred from endpoint validation metrics alone.
Based on the trajectory-level precursors we identify, we propose
the \textbf{MTR (Monitor–Trust–Regulator)} framework, illustrated
in Figure~\ref{fig:mtr_framework}. MTR operates a secondary
trajectory-level control loop alongside the base learner:
it monitors trajectory statistics, aggregates them over a
slower timescale into a trust estimate $\tau$, and adaptively
modulates the effective learning intensity. MTR requires no
continued access to pristine real data beyond a fixed anchor set
— only a small, fixed set of anchor examples drawn from the initial
checkpoint. We validate MTR on recursive language modeling (TinyStories)
and recursive pseudo-labeling on CIFAR-10 under a stress-test regime.
In both settings, MTR detects collapse precursors before visible
degradation emerges and can stabilize learning.

Our contribution is not the observation that recursive models
can collapse, but the identification of trajectory-level
precursors that reveal collapse before visible failure and
enable intervention. This work reframes the study of iterative
learning degradation in terms of trajectory-level observability
of recursive collapse. MTR serves as one demonstration that
the precursor regime is actionable.

\section{What is silent collapse?}

\paragraph{Definition.}
We define \textit{silent collapse} as a dynamical regime in which
iterative learning systems undergo progressive internal degeneration
while endpoint validation metrics remain stable or improve.
Internal degeneration refers to contraction of predictive entropy,
loss of representational diversity, and in some settings erosion of
support statistics, coupled with a progressive loss of recoverability.
Endpoint validation metrics include loss, perplexity, and accuracy.
Silent collapse therefore reflects a decoupling between visible
predictive performance and internal dynamical health.

\paragraph{Trajectory-Level Precursors.}
We define \textit{trajectory-level precursors} as reproducible changes
in internal trajectory statistics that systematically emerge before
visible degradation in endpoint performance metrics.  In this work,
we focus on two robust precursors: (1) contraction of anchor entropy,
and (2) contraction of representation drift.  (In some settings,
erosion of support statistics provides additional evidence.)  These
precursors are measurable, exhibit a consistent onset order relative
to visible collapse, and are coupled with a progressive loss of
recoverability.

\paragraph{Early and late silent collapse.}
Silent collapse follows a consistent dynamical sequence.
\textit{Early silent collapse} corresponds to the hidden contraction
precursor regime: predictive entropy and representation drift contract
(and in some settings, support statistics contract) while endpoint
validation metrics remain stable or improve. In this regime, the system
is already degrading but conventional evaluation remains deceptively stable.
\textit{Late silent collapse} corresponds to visible degradation:
perplexity explosion, accuracy loss, calibration collapse, or support
erosion become detectable. This progression defines a critical
early-warning window during which intervention can substantially
mitigate later collapse.

\paragraph{Sources of silent collapse.}
Silent collapse arises from three coupled sources:

\textit{Feedback reliability error.} The feedback used for learning
or inference cannot be locally verified. Examples include synthetic
text, pseudo-labels, biased reward signals \citep{leike2018scalable},
and model-derived scientific quantities used in downstream inference
pipelines. Each individual signal may appear locally plausible, yet
the distribution from which it is drawn may be systematically distorted.

\textit{Trajectory amplification error.} Small reliable-looking
biases are repeatedly reinforced across iterations. A mild
concentration of probability mass in one generation becomes amplified
in the next, progressively eroding support diversity, reducing
predictive entropy, and contracting representation dynamics.

\textit{Endpoint observability error.} Endpoint validation metrics
compress high-dimensional learning dynamics into low-order aggregate
statistics \citep{vapnik1999nature}. As a result, substantially
different internal dynamical states — including healthy learning
and silent collapse — may produce observationally indistinguishable
endpoint behavior.

This defines an endpoint-metric non-identifiability problem:
distinct internal dynamical states cannot always be uniquely
distinguished from endpoint validation behavior alone.
Together, these mechanisms define silent collapse as a structural
failure of observability in iterative learning systems.

\section{Theoretical intuition: hidden contraction precedes visible collapse}

Hidden contraction is the early precursor regime of silent collapse.
It is the regime in which internal dynamical health deteriorates
before endpoint metrics become visibly degraded.

We begin by establishing a conceptual framework for understanding silent collapse.
Standard validation metrics—perplexity, loss, accuracy—are low-order aggregate statistics.
They average performance over many examples and can remain stable even when the support
of the internal predictive distribution is already contracting.

\begin{center}
\fbox{\parbox{0.9\linewidth}{\centering
\textbf{Proposition 1 (Hidden contraction can precede visible collapse).} \\
In recursive learning, collapse can first manifest as a progressive contraction
of trajectory-level observables—predictive entropy and representation drift—
before any detectable degradation in standard validation metrics. This motivates
the notion of a \textit{hidden contraction regime} in which the system is already
degrading but conventional evaluation remains deceptively stable.
}}
\end{center}

\subsection*{Intuition}

Validation perplexity and loss are dominated by the most probable modes of the distribution.
Early iterations still preserve these dominant modes, so short-term predictive
performance remains stable. However, support diversity and inter-generational
representational variability contract much earlier. Iterative learning progressively
concentrates probability mass into dominant modes while eroding low-support regions of
the distribution. As a result, apparent short-term predictive quality can remain stable
despite progressive loss of support diversity. This phenomenon bears conceptual
similarity to habitat loss and biodiversity decline in ecological systems,
where aggregate measures can mask the progressive erosion of rare species
\citep{hanski2011habitat}. Trajectory-level observables—predictive
entropy and representation drift—are sensitive to this concentration
process because they monitor the evolution of the learning dynamics themselves, rather
than only endpoint prediction quality. As iterative optimization pressure accumulates,
the system eventually exits the hidden contraction regime and enters visible collapse.

\subsection*{Why endpoint metrics fail}

Endpoint metrics fail because they average over the most probable modes of the distribution.
Early in recursive training, these dominant modes remain intact, so short-term predictive
performance stays stable. However, low-support regions of the distribution — where
degradation first appears — have negligible contribution to aggregate metrics such as
perplexity or accuracy. This reflects a broader statistical principle: low-order
aggregate statistics cannot uniquely characterize high-dimensional distributions
\citep{casella2002statistical}. Trajectory-level observables avoid this averaging
because they monitor the geometry of the learning process itself: entropy measures
uncertainty across the full distribution; representation drift tracks changes in
internal representations; tail coverage directly measures low-support retention.
These statistics reveal hidden contraction before endpoint metrics show any change.

\subsection*{Theoretical consequence}

If endpoint metrics are dominated by high-probability modes,
while low-support regions progressively contract under iterative
feedback, then multiple internally distinct learning states may
remain observationally indistinguishable under endpoint validation.
Consequently, endpoint validation behavior may not reliably determine
internal dynamical state. This defines the fundamental limitation
underlying silent collapse: standard evaluation cannot reliably infer
internal dynamical health from endpoint performance alone.

\begin{center}
\fbox{\parbox{0.9\linewidth}{\centering
\textbf{Proposition 2.} \\
Under iterative learning from unverifiable feedback,
endpoint validation behavior may not reliably determine
internal dynamical state.
}}
\end{center}

\subsection*{Trajectory contraction statistic}

To quantify hidden contraction, we define a statistic that
measures inter-generational representational drift on a fixed anchor set.
Representations are mean-pooled over the anchor set before computing the statistic;
mean pooling provides a stable low-dimensional summary of inter-generational
representation dynamics. Drift is measured as the normalized squared Euclidean
distance between consecutive mean-pooled representations:

\[
S_g = \frac{1}{W} \sum_{k=0}^{W-1} \frac{\| \mathbf{z}_{g-k} - \mathbf{z}_{g-k-1} \|^2}{\| \mathbf{z}_{g-k-1} \|^2 + \epsilon},
\]

where $\mathbf{z}_g$ denotes the model's penultimate-layer representations (mean-pooled
over the anchor set) at generation $g$, $W$ is a smoothing window, and $\epsilon$ is a
small constant for numerical stability. Under hidden contraction, support contraction
reduces effective trajectory variability, leading to progressively smaller $S_g$.

\subsection*{The hidden contraction regime}

We formally define the hidden contraction regime as the set of iterations where
trajectory-level observables have significantly degraded but standard metrics have
not yet shown deterioration. For example, we empirically observe:

\[
\frac{H_g}{H_0} < 0.5 \quad \text{while} \quad \frac{\text{PPL}_g}{\text{PPL}_0} \approx 1,
\]

where $H_g$ is anchor entropy and $\text{PPL}_g$ is validation perplexity.
During this regime, standard evaluation incorrectly suggests stable learning dynamics
despite ongoing internal collapse. This regime can persist for multiple
iterations before visible validation degradation emerges, establishing a window for
early intervention.

Hidden contraction is the observable precursor manifestation of silent collapse dynamics.
We therefore view hidden contraction as a recurring dynamical phase of iterative
learning systems, reflecting a dynamical transition toward contracted dynamics
under self-reinforcing feedback.

\subsection*{Implication for regulation}

Because collapse can first manifest in trajectory-level contraction,
regulation mechanisms that operate on trajectory-level observables can intervene
before visible degradation emerges. Trajectory statistics evolve earlier than
endpoint metrics; therefore regulation based on trajectory state can intervene
before visible failure occurs. MTR provides one concrete implementation of
trajectory-level regulation.

\begin{figure}[t]
\centering
\begin{tikzpicture}
\draw[thick, ->] (0,0) -- (10,0);
\node at (0,0.3) {Healthy learning};
\node at (3.5,0.3) {Hidden contraction regime};
\node at (8,0.3) {Visible degradation};
\draw[decorate,decoration={brace,amplitude=8pt,mirror}] (1.5,-0.8) -- (6.5,-0.8);
\node[align=center] at (4,-1.2) {Trajectory contraction \\ without visible failure};
\fill[red!20] (1.5,-0.2) rectangle (6.5,0);
\fill[red!40] (6.5,-0.2) rectangle (9.5,0);
\node at (8.5,-0.5) {Metric failure};
\end{tikzpicture}
\caption{Hidden contraction reliably precedes visible collapse across recursive trajectories,
providing a measurable early-warning window.}
\label{fig:hidden_contraction_regime}
\end{figure}

\section{Results}

We evaluated silent collapse across two fundamentally different iterative
learning modalities: language generation on TinyStories (main results) and
recursive pseudo-labeling on CIFAR-10 under a stress-test regime (cross-modal
validation).  All quantitative results reported in this section are based on
the TinyStories dataset unless stated otherwise.

\paragraph{Result 1: Trajectory-level precursors precede visible collapse.}
To quantify the temporal structure of silent collapse, we analysed five
independent open‑loop recursive training runs (5 seeds) on TinyStories.
For each seed we defined:

\begin{itemize}
    \item \textbf{Hidden contraction onset}: the first generation where
          anchor entropy \(H_g/H_0 < 0.5\) (or, when entropy alone was ambiguous,
          the minimum of entropy and drift criteria).
    \item \textbf{Visible collapse onset}: the first generation where
          validation perplexity exceeds \(5\times\) the minimum perplexity
          achieved up to that generation and never later recovers below that
          threshold (irreversible collapse).
\end{itemize}

Across the five seeds, the hidden contraction onset had a median of
generation 4 (interquartile range 4–6), whereas the visible collapse onset
had a median of generation 6 (interquartile range 6–6).  Hidden contraction
preceded visible collapse in all five seeds (5/5) and the median lead time
was 2 generations.  Thus, collapse is not abrupt; it contains a measurable,
reproducible trajectory‑level precursor regime.

\paragraph{Result 2: Hidden contraction is accompanied by progressive loss of recoverability.}
From the same open‑loop trajectory (seed 0) we extracted checkpoints at
generations 2, 4, 6 and 8 and performed fixed‑budget recovery experiments.
Under a weak intervention budget (0.5\% real data, 400 training steps,
i.e. one quarter of a normal generation), recovery validation perplexity
was substantially higher for the late checkpoints: 63.9 (g2), 57.3 (g4),
92.0 (g6) and 123.9 (g8).  A strong intervention (10\% real data, full 1600 steps)
restored all checkpoints to similar perplexity ($\approx$ 37–41).  Table~\ref{tab:recoverability}
summarizes the results for all three intervention strengths.

\begin{table}[t]
\centering
\caption{Recovery validation perplexity under fixed budgets (seed 0).}
\begin{tabular}{cccc}
\hline
Checkpoint & Weak (0.5\% real, 400 steps) & Medium (2\% real, 800 steps) & Strong (10\% real, 1600 steps) \\
\hline
g2 & 63.93 & 51.25 & 41.40 \\
g4 & 57.28 & 45.17 & 36.85 \\
g6 & 91.96 & 53.51 & 37.57 \\
g8 & 123.92 & 61.29 & 38.81 \\
\hline
\end{tabular}
\label{tab:recoverability}
\end{table}

Thus, recoverability weakens before catastrophic endpoint degradation becomes
visible, indicating that internal learning dynamics have already contracted.

\paragraph{Result 3: Control baselines for recursive pressure.}
For completeness, we compare open‑loop, MTR, a fixed low‑pressure baseline
($\alpha$=0.5) and a randomly shuffled trust schedule (random $\tau$) in Table~\ref{tab:controls}.
Open‑loop collapses in all runs, while all three regulated conditions prevent
visible collapse.  However, the fixed low‑pressure baseline uses a much higher
average mixing ratio (0.5) and therefore does not provide a fair comparison
in terms of synthetic exposure.  We therefore focus on a same‑pressure
comparison in the next result.

\begin{table}[t]
\centering
\caption{Control baselines (open‑loop, MTR, fixed‑low‑alpha, random‑$\tau$).}
\begin{tabular}{lcccc}
\hline
\textbf{Baseline} & \textbf{Final PPL} & \textbf{Final $H$} & \textbf{Mean $\alpha$} & \textbf{Collapse frac.} \\
\hline
open\_loop & $1.89\times10^8 \pm 3.21\times10^8$ & $0.317 \pm 0.021$ & $0.893 \pm 0.000$ & $1.00$ \\
MTR (entropy‑regulated) & $11.91 \pm 0.10$ & $2.104 \pm 0.037$ & $0.270 \pm 0.011$ & $0.00$ \\
fixed\_low\_alpha ($\alpha=0.5$) & $13.72 \pm 0.09$ & $2.140 \pm 0.057$ & $0.500 \pm 0.000$ & $0.00$ \\
random\_$\tau$ & $11.56 \pm 1.13$ & $2.057 \pm 0.151$ & $0.286 \pm 0.020$ & $0.00$ \\
\hline
\end{tabular}
\label{tab:controls}
\end{table}

\paragraph{Result 4: Same‑pressure comparison: MTR vs fixed $\alpha$=0.27.}
To fairly compare MTR with a constant‑pressure baseline that uses the same
average synthetic exposure, we ran a fixed‑mixing experiment with constant
\(\alpha = 0.27\) (the mean effective \(\alpha\) of MTR).  Table~\ref{tab:same_pressure}
shows the results over five seeds.

\begin{table}[t]
\centering
\caption{Same‑pressure comparison: MTR vs fixed $\alpha$=0.27.}
\begin{tabular}{lccc}
\hline
\textbf{Method} & \textbf{Final PPL} & \textbf{Final $H$} & \textbf{Collapse fraction} \\
\hline
MTR (entropy‑regulated) & $11.91 \pm 0.10$ & $2.103 \pm 0.037$ & $0/5$ \\
Fixed $\alpha=0.27$ & $10.94 \pm 0.21$ & $2.105 \pm 0.043$ & $0/5$ \\
\hline
\end{tabular}
\label{tab:same_pressure}
\end{table}

The fixed baseline achieves slightly better final perplexity, but it was
constructed \emph{retrospectively} using the observed mean pressure of MTR.
In practice, such retrospective knowledge is not available a priori.  The
TinyStories environment is persistently high‑risk, where a conservative
constant pressure works well; under non‑stationary or unknown reliability
regimes, adaptive regulation (MTR) would be advantageous.  Thus, MTR
provides an online adaptive strategy rather than a universally superior
constant‑pressure substitute.

\begin{figure*}[t]
\centering
\includegraphics[width=\textwidth]{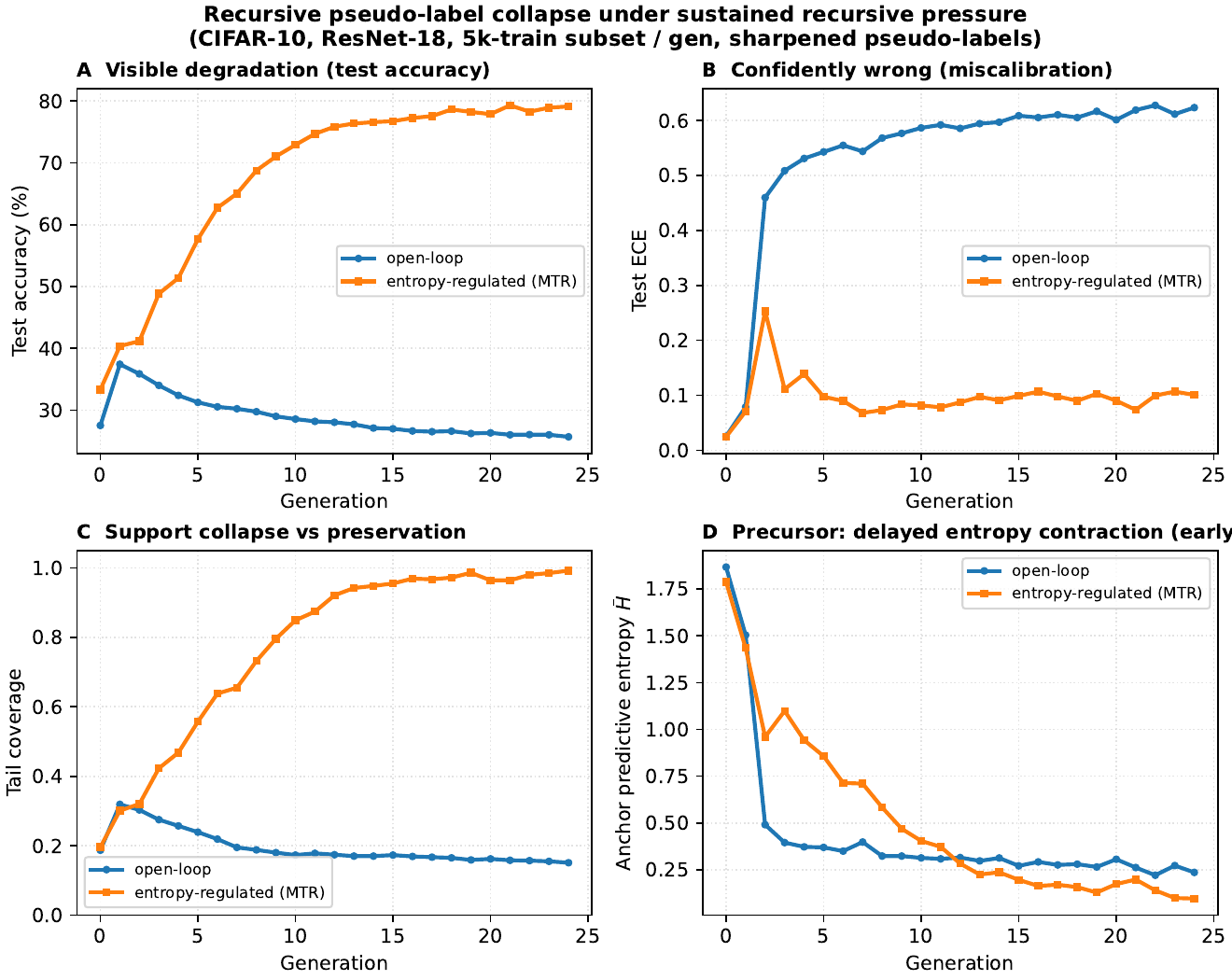}
\caption{\textbf{Recursive pseudo-label collapse under sustained pressure.}
Open-loop exhibits visible degradation and calibration collapse;
MTR mitigates these failures and improves stability under sustained recursive pressure.}
\label{fig:sl_recursive_pressure}
\end{figure*}

\paragraph{Result 5: Hidden contraction generalizes across learning modalities.}
We tested the same early‑to‑late sequence on CIFAR‑10 under a stress‑test
regime (25 generations, 5 k training subset, sharpened pseudo‑labels).
Open‑loop accuracy progressively declined after an initial peak, while
MTR consistently improved stability relative to open‑loop.  Entropy
contraction reliably preceded visible degradation, with support erosion
and calibration failure emerging later in many runs (Fig.~\ref{fig:sl_recursive_pressure}).

\subsection{Scaling to a standard language benchmark: GPT‑2 on WikiText‑2}

To determine whether silent collapse generalizes to a larger, more realistic
model, we repeated the recursive training on GPT‑2 (124 M parameters) with
WikiText‑2.  Table~\ref{tab:gpt2_results} reports anchor entropy \(H_g\),
validation perplexity (PPL), and trust variable \(\tau\) for generations 0–10.

\begin{table}[t]
\centering
\caption{GPT‑2 results: open-loop, 10\% real mixing, and MTR (entropy‑regulated).}
\begin{tabular}{c|ccc}
\hline
\textbf{Gen} & \textbf{open\_loop} & \textbf{open\_loop\_10real} & \textbf{entropy\_regulated (MTR)} \\
\hline
0 & H=3.48, PPL=30.9 & same as left & same as left \\
1 & H=3.07, PPL=25.7 & H=2.15, PPL=27.8 & H=3.07, PPL=25.7 \\
2 & H=2.64, PPL=23.0 & H=1.69, PPL=32.9 & H=2.80, PPL=23.4, $\tau=0.78$ \\
3 & H=2.12, PPL=25.2 & H=1.71, PPL=34.0 & H=2.63, PPL=21.9, $\tau=0.64$ \\
4 & H=1.25, PPL=50.1 & H=1.79, PPL=33.8 & H=2.51, PPL=21.5, $\tau=0.57$ \\
5 & H=1.05, PPL=74.7 & H=1.79, PPL=33.6 & H=2.56, PPL=20.9, $\tau=0.52$ \\
6 & H=0.94, PPL=102.5 & H=1.87, PPL=33.3 & H=2.56, PPL=20.5, $\tau=0.54$ \\
7 & H=0.87, PPL=130.0 & H=1.98, PPL=31.9 & H=2.55, PPL=20.2, $\tau=0.54$ \\
8 & H=0.83, PPL=156.4 & H=2.05, PPL=30.8 & H=2.54, PPL=20.0, $\tau=0.53$ \\
9 & H=0.80, PPL=181.5 & H=2.10, PPL=30.4 & H=2.52, PPL=20.5, $\tau=0.53$ \\
10& H=0.78, PPL=208.0 & H=2.21, PPL=29.7 & H=2.46, PPL=21.0, $\tau=0.52$ \\
\hline
\end{tabular}
\label{tab:gpt2_results}
\end{table}

Open‑loop shows a clear silent‑collapse precursor (entropy decline before
perplexity rise), while MTR maintains stable entropy and low perplexity,
confirming the phenomenon is not restricted to TinyStories.

\subsection{Summary of results}

Across language generation and image classification, hidden contraction
precedes visible degradation.  Standard validation metrics remained stable
initially, yet trajectory-level precursors provided early warning.  Silent
collapse is therefore predictable and partially preventable, and trajectory‑
aware regulation provides one adaptive method without requiring prior
knowledge of an optimal constant pressure.

\section{Discussion}

Silent collapse is not specific to language generation or recursive self‑training.
It emerges whenever iterative learning systems optimize against feedback whose
reliability cannot be verified locally.  This includes RLHF, federated learning,
reinforcement learning with reward hacking, continual learning, and multi‑agent
systems.  In all these settings, standard validation metrics may remain
deceptively stable while internal dynamics progressively contract.  This is
not a practical nuisance but a structural limitation: endpoint metrics
collapse high‑dimensional dynamics into low‑order statistics and cannot
reliably infer internal health.

\subsection*{Adaptive regulation under changing reliability regimes}
The same‑pressure comparison shows that a fixed mixing ratio
\(\alpha = 0.27\) – chosen after the fact to match MTR’s average
pressure – achieves slightly better final perplexity than MTR on
the static, high‑risk TinyStories environment.  However, this
baseline assumes retrospective knowledge of a near-optimal constant
pressure, which is not available when the reliability of the learning
environment (e.g., data quality, task distribution, or feedback
noise) changes over time.  Under non‑stationary conditions, a
fixed pressure may become suboptimal or even dangerous, whereas
trajectory‑aware regulation can adapt online by monitoring entropy
and representation dynamics.  This adaptive capability is the main
value of MTR: it provides a principled, data‑driven way to modulate
recursive pressure in unknown or changing environments.

\subsection*{Hierarchy of findings}
To clarify the contribution, we distinguish three levels:
\begin{itemize}
    \item \textbf{Phenomenon}: silent collapse – a regime where internal
          contraction and loss of recoverability precede visible degradation.
    \item \textbf{Precursor dynamics}: hidden contraction (entropy and
          representation drift) with reproducible onset ordering.
    \item \textbf{Intervention}: MTR illustrates that the hidden phase
          provides an opportunity for early intervention; it is not the
          only way, but one adaptive strategy.
\end{itemize}

\subsection*{Recoverability as evidence of contracted dynamics}
The progressive loss of recoverability suggests that silent collapse is
not merely a transient statistical fluctuation, but reflects a genuine
contraction of the accessible learning dynamics.  This is particularly
evident in the weak‑intervention recovery experiments: once the system
has progressed past the hidden‑to‑visible transition, restoring healthy
behavior requires substantially more real data.

We have identified silent collapse as a predictable dynamical sequence:
early hidden contraction (entropy contraction, representation drift contraction)
precedes visible degradation.  These precursors enable early warning,
manifesting before visible changes in standard metrics.  The hidden contraction
phase provides an opportunity for early intervention, and MTR demonstrates
that such intervention can be realized without requiring retrospective tuning
of a constant pressure regime.

Limitations include the need for a fixed anchor set (2000 examples sufficed)
and modest computational overhead.  Future work could explore dynamic anchor
updating, integration with differential privacy, and deeper geometric
characterization of representation collapse (e.g., effective rank or spectral
analysis).

Recursive learning systems may appear healthy under conventional validation
while already undergoing substantial trajectory contraction.  Standard
validation metrics are systematically insufficient to guarantee model health;
visible validation stability should no longer be interpreted as evidence of
stable learning dynamics.  Reliable monitoring requires trajectory‑level
information that is not reliably recoverable from endpoint metrics alone.
Distinct internal dynamical states may therefore remain observationally
indistinguishable under standard validation metrics alone.

\section{Methods}

\subsection{Recursive language modeling on TinyStories}

We used the TinyStories dataset.  The base model was a compact autoregressive
transformer.  Each generation was trained for 2 epochs with learning rate
5e‑5 and batch size 64.  The anchor set comprised 2,000 fixed prompts.
Synthetic data was generated with nucleus sampling (top‑p=0.9, temperature=0.7).
For entropy‑regulated training, the trust variable was computed as
\(\tau_g = \max(0.2, \min(1.0, (H_{g-1}/H_0)^2))\).  The planned mixing schedule
was \(\alpha_{\text{planned}} = [0, 0.25, 0.5, 0.75, 1.0, 1.0, \ldots]\).

\subsection{Fixed‑pressure baseline (constant \(\alpha=0.27\))}

We ran an additional baseline where the mixing proportion was fixed to
\(\alpha = 0.27\) for all generations \(g \ge 1\) (generation 0 used real
data with \(\alpha=0\)).  This value was chosen retrospectively to match the
mean effective pressure of MTR across the five seeds.  All other hyperparameters
were identical to the MTR condition.  This baseline thus assumes retrospective
knowledge of a near-optimal constant pressure regime and is not a practical prior.

\subsection{Recovery experiments}

Recovery experiments used the same open‑loop trajectory (seed 0).  From each
checkpoint we fine‑tuned the model for a fixed number of steps (400, 800 or
1600) on a mixture of the frozen synthetic pool (Pool‑B) and a small fraction
of real data.  The real fractions were 0.5\%, 2\% and 10\%.

\subsection{Recursive pseudo‑labeling on CIFAR‑10 under sustained pressure}

We used CIFAR‑10 with ResNet‑18 adapted for 32×32 inputs.  A fixed subset of
5,000 training images was used.  Generation 0 used ground‑truth labels (10 epochs,
lr 0.1, milestones at epochs 5 and 8).  Generation 1 used 50\% synthetic, and
generation 2 onward used fully synthetic supervision.  Pseudo‑labels were
sharpened with temperature 0.1.  The recursive loop ran for 25 generations.

\subsection{Evaluation metrics}

Anchor entropy \(H_g\): average predictive entropy over the anchor set.
Representation drift \(S_g\): normalized squared Euclidean distance between
penultimate‑layer outputs on the anchor set across consecutive generations,
as defined in the main text.
Rare‑token mass (language): proportion of tokens whose types are below the
10th percentile of training frequency.
Tail coverage (classification): proportion of low‑support classes with
prediction probability above a threshold.
Expected calibration error (ECE): computed with 15 bins.

\subsection{Code and data availability}

All code and data are available at \url{https://github.com/...}
(repository will be made public upon publication).

\begin{figure*}[t]
\centering
\includegraphics[width=0.7\textwidth]{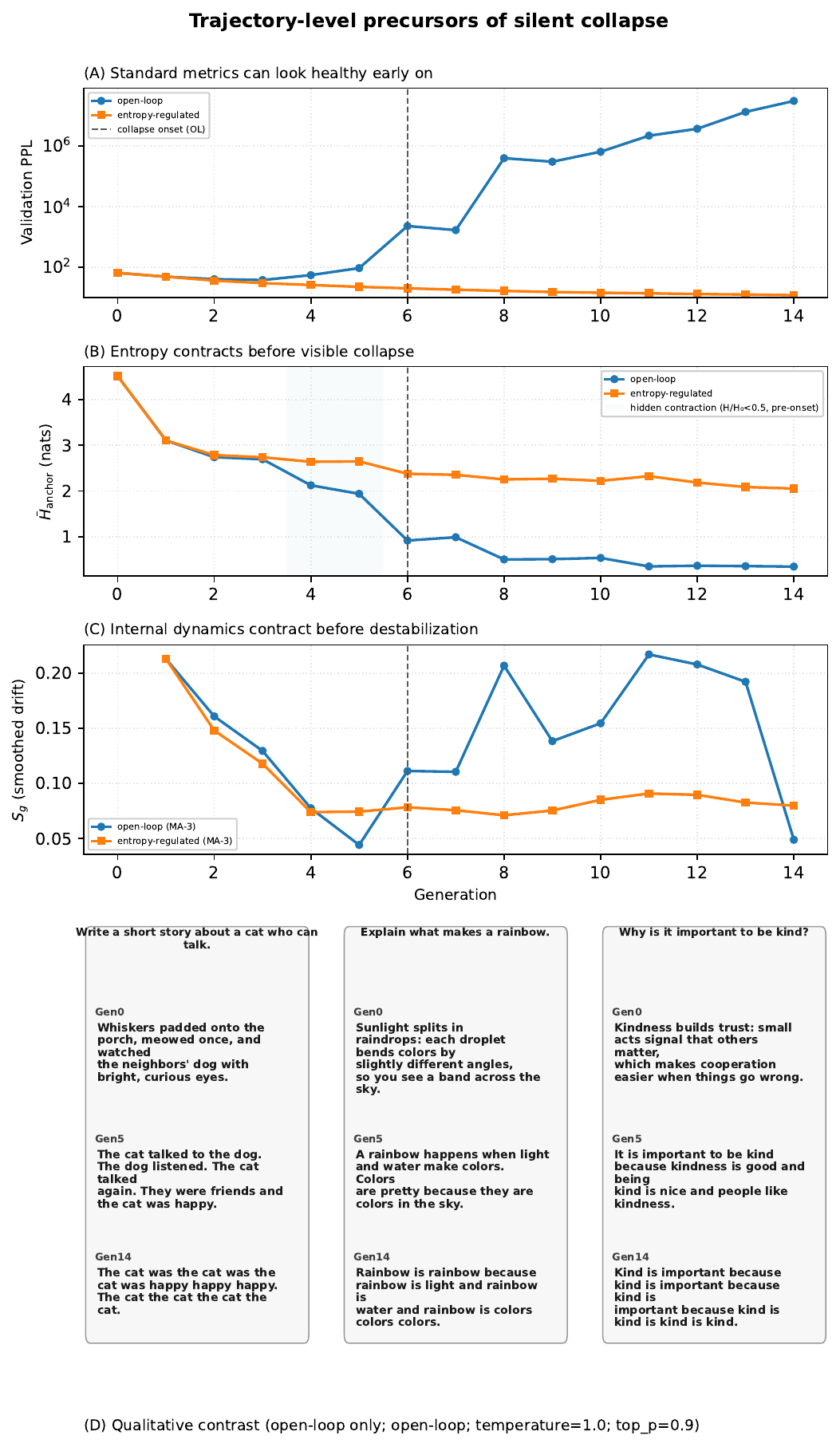}
\caption{\textbf{Hidden contraction reliably precedes visible collapse across recursive trajectories.}
(A) Standard validation metrics initially remain stable.
(B) Predictive entropy contracts substantially earlier than visible collapse.
(C) Inter-generational representation drift reveals hidden dynamical changes.
(D) Qualitative generation examples (if present).}
\label{fig:trajectory_precursors}
\end{figure*}

\begin{figure*}[t]
\centering
\includegraphics[width=\textwidth]{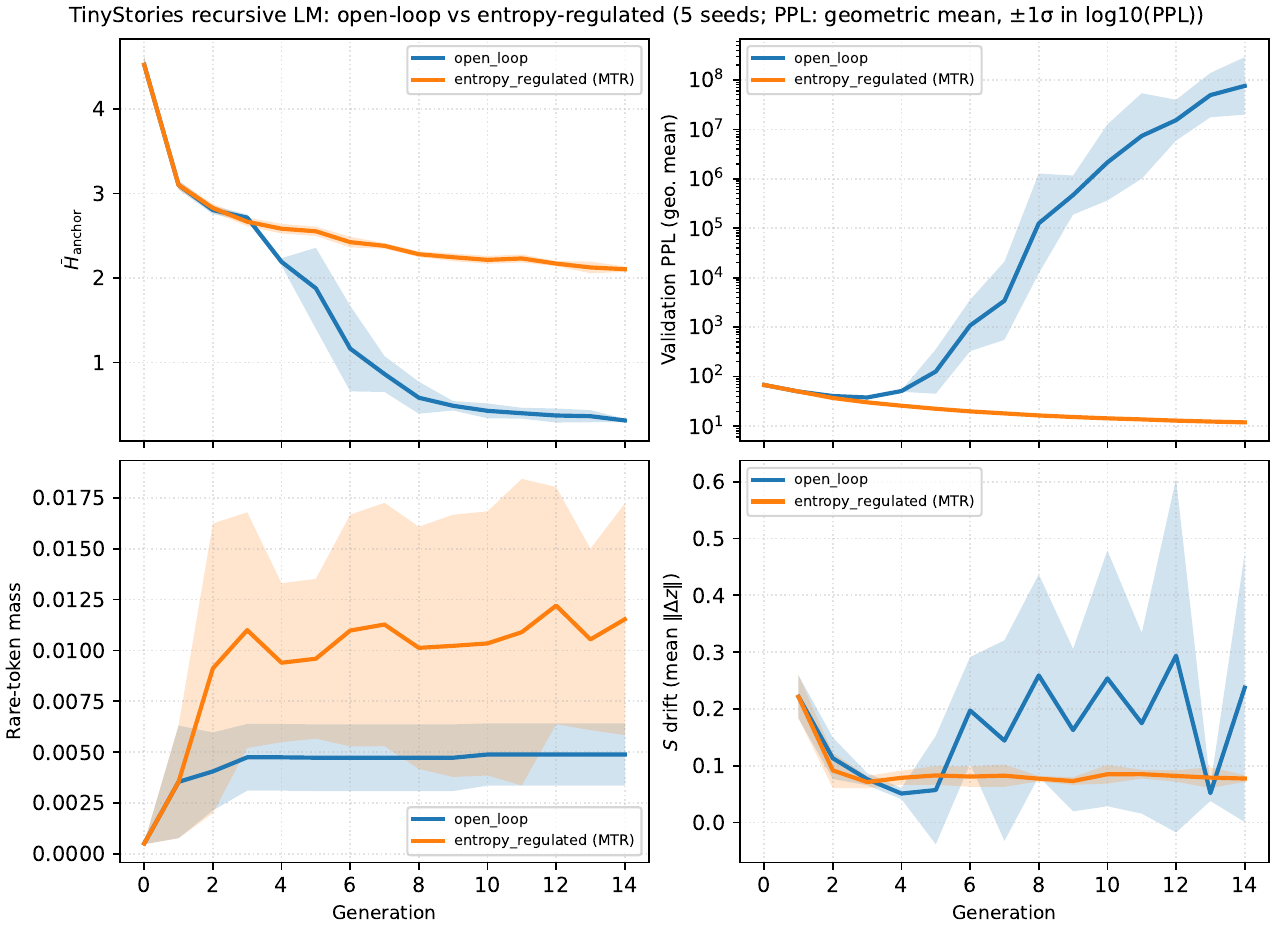}
\caption{\textbf{Multi-seed robustness.}
Open-loop recursive training exhibits catastrophic late-stage degradation across all seeds,
whereas entropy-regulated training preserves stable entropy, bounded perplexity,
and stable inter-generational dynamics.}
\label{fig:rlm_multiseed}
\end{figure*}

% ...


\bibliographystyle{unsrtnat}
\bibliography{references}
\end{document}